\documentclass[letterpaper, 10 pt, conference]{ieeeconf}  

\IEEEoverridecommandlockouts                              

\usepackage{amsmath, amsfonts, amssymb}
\usepackage{graphicx}
\usepackage{xcolor}
\usepackage{float}
\usepackage{cite}
\usepackage{bm}
\usepackage{svg}
\usepackage{algorithm}
\usepackage{algpseudocode}

\usepackage{booktabs}
\usepackage{multirow}

\usepackage{hyperref}  

\newif\ifanonymous
\anonymousfalse           

\title{\LARGE \bf
Agile Monoped Hopping on Complex and Discrete Terrain via Kinodynamic Planning with Distilled MPPI Steering
}

\ifanonymous
\author{Anonymous Authors}
\else
\author{Yue Qin, Denglin Cheng, Jingyu Liu, and Yanran Ding
\thanks{Authors are with the Department of Robotics, University of Michigan, Ann Arbor, MI 48109, USA.
        {\tt\small \{yueqin, denglinc, juliusl, yanrand\}@umich.edu}}%
}
\fi

\begin{document}

\algrenewcommand\algorithmicrequire{\textbf{Input:}}
\algrenewcommand\algorithmicensure{\textbf{Output:}}

\maketitle
\begin{abstract}
This paper presents FlowRRT, a kinodynamic planner for agile monoped hopping on discrete terrain. Steering for kinodynamic RRT on complex dynamical systems currently relies on motion primitives or closed-loop policies, limiting the diversity of motions or the planning speed. Distilling a trajectory optimizer avoids both, but existing methods regress a single solution per query, which fails on monoped hopping, where the feasible action set is narrow and disconnected. Our key insight is to treat steering as set-valued and predict the distribution of feasible actions instead. A multi-start MPPI expert collects the entire feasible set of each query offline, and conditional flow matching distills it into a generative sampler. With this sampler, FlowRRT coordinates successive jumps over long horizons and finds non-greedy maneuvers that are hard for reactive policies to discover. Benchmarks across different terrains show higher success rates and substantially shorter planning times than sampling-based baselines. On terrains with inclined platforms, FlowRRT reaches $80\%$ success within $1$~s, versus $24$~s for RL-RRT and $77$~s for SST. Sim-to-sim tracking in MuJoCo confirms accurate execution under full-order dynamics.
\end{abstract}

\section{Introduction}
\label{sec:intro}

Monoped robots possess substantial potential to navigate unstructured environments owing to their lightweight morphology and dynamic hopping gait~\cite{salto2017,mppc2024}, with applications such as disaster rescue and planetary exploration.
Compared with quadruped and bipedal platforms, monoped hopping faces the unique challenge of severe underactuation.
Selecting appropriate footholds is therefore critical, as each foothold must regulate the robot's momentum to avoid collisions and prepare for subsequent jumps. 
The combination of highly underactuated dynamics and discrete foothold decisions makes long-horizon kinodynamic planning for monoped hopping particularly challenging and underexplored.

\begin{figure}[tb]
       \centering
       \includegraphics[width=\columnwidth]{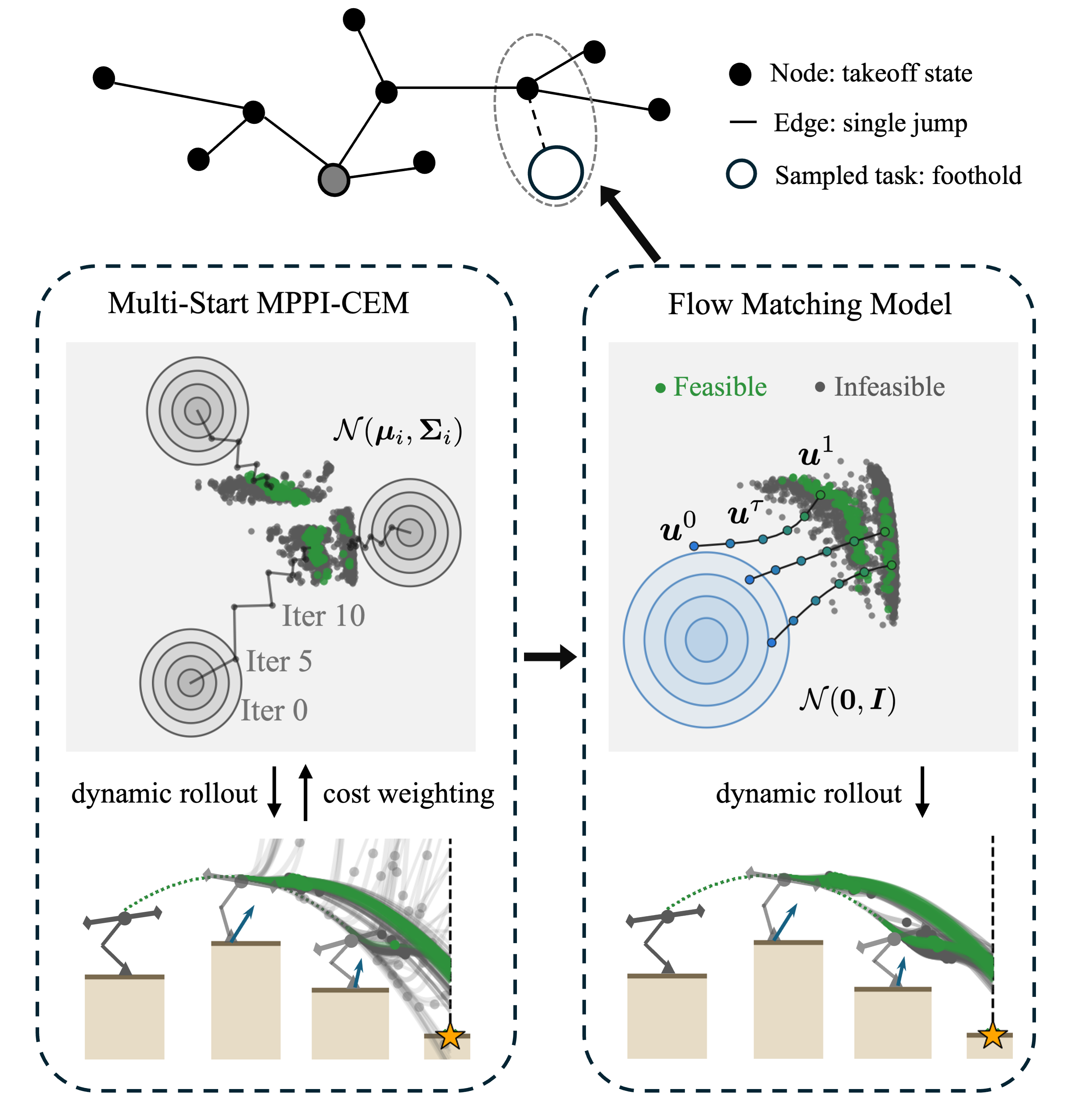}
\caption{Overview of the distilled MPPI steering. Left, the multi-start MPPI-CEM expert solves one steering query offline and collects every feasible action it encounters, which form several distinct clusters. Right, a conditional flow matching model trained on these actions maps Gaussian noise to the same clusters in a single online query, preserving the multimodality of the expert data.}
       \label{fig:method_overview}
\end{figure}

Prior work has tackled this problem with mixed-integer programming (MIP)~\cite{mppc2024,posa2025perceptive} and reinforcement learning (RL)~\cite{zheng2025jump}, which jointly optimize control and footholds as a monolithic formulation. However, MIP scales poorly with the number of jumps, and the reactive nature of RL limits long-horizon reasoning.
In contrast, tree-based motion planners such as rapidly exploring random trees (RRT)~\cite{lavalle2001randomized} offer a distinct advantage, as they decouple long-horizon foothold planning from successive local steering~\cite{shkolnik2011bounding,aaron2020fast,ding2021hybrid}.
The bottleneck of this approach lies in the steering function, as each tree expansion requires solving a two-point boundary value problem (BVP) under dynamic constraints, which is computationally expensive.

Existing methods address this challenge mainly in three ways: motion primitives, closed-loop steering, and optimizer distillation. Primitive-based methods restrict steering to a precomputed library or a structured subset of motions under simplifying assumptions~\cite{shkolnik2011bounding,ding2021hybrid}, while closed-loop steering drives the system toward the target with a goal-conditioned policy~\cite{rlrrt2019,hassidof2025trainonce,atreya2022s3f,li2021mpcmpnet}. Both make steering tractable, but primitives limit the diversity of motions, whereas closed-loop steering requires a policy query at every step of forward propagation, which remains costly.
Optimizer distillation trains a model to imitate a trajectory optimizer, which avoids both limitations above by directly generating the whole action trajectory for each steering query~\cite{wolfslag2018rrtcolearn}.
However, existing methods mostly regress a single solution per query, which can average the distinct solutions of nearby queries into an infeasible one~\cite{wolfslag2018rrtcolearn} and discard the alternative solutions that are valuable for tree expansion. These drawbacks have limited the application of this approach to more complex systems such as monoped hopping, where contact and discrete footholds make such distinct solutions both common and indispensable.

In this paper, we propose FlowRRT, a kinodynamic RRT algorithm that efficiently generates long-horizon, dynamically feasible motion trajectories for monoped hopping on complex and discrete terrain. Our key insight is that, for legged systems with contact on discrete terrain, the feasible region of a steering query is typically narrow and disconnected, so the steering function should predict the distribution of feasible actions rather than a single solution. Fig.~\ref{fig:method_overview} illustrates this idea, where a multi-start model predictive path integral (MPPI) expert collects the feasible actions of each query offline and a distilled conditional flow matching (CFM) sampler generates them in a single query online. Combined with task-space sampling and a learned reachability estimator, FlowRRT solves $93\%$ of the planning problems within $1$~s in simulation, where SST~\cite{li2016asymptotically} and RL-RRT~\cite{rlrrt2019} solve $26\%$ and none.

\section{Dynamics Model}
\label{sec:modeling}
\subsection{Robot Model}
The monoped robot is composed of a floating torso and a two-degree-of-freedom leg with a point foot. To reduce the computational load, a single-rigid-body (SRB) model is used to describe the robot's dynamics, which is justified since the hip and knee motors are mounted on the torso and the leg links account for less than $10\%$ of the total mass.
The robot state $\bm{x} \in \mathbb{R}^6$ comprises its center of mass (CoM) position $\bm{p}_{\mathrm{c}} \in \mathbb{R}^2$, torso pitch $\theta \in \mathbb{R}$, and their time derivatives.
Given the mass $m$ and inertia $I$ of the robot, the dynamics can be expressed as:
\begin{equation}
    \dot{\bm{x}}
    = \frac{\mathrm{d}}{\mathrm{d}t}
    \begin{bmatrix}
        \bm{p}_{\mathrm{c}} \\
        \theta \\
        \dot{\bm{p}}_{\mathrm{c}} \\
        \dot{\theta}
    \end{bmatrix}
    = \begin{bmatrix}
        \dot{\bm{p}}_{\mathrm{c}} \\
        \dot{\theta} \\
        \bm{\lambda}/m + \bm{g} \\
        (\bm{r} \wedge \bm{\lambda})/I
    \end{bmatrix},
    \label{eq:srb-dynamics}
\end{equation}
where $\bm g$ is the gravitational acceleration, $\bm{\lambda} \in \mathbb{R}^2$ is the ground reaction force (GRF) applied at the stance foot, and $\bm{r} = \bm p_{\mathrm{f}}^{\mathrm{st}} - \bm p_{\mathrm{c}}$ is its moment arm, with $\bm p_{\mathrm{c}}, \bm p_{\mathrm{f}}^{\mathrm{st}} \in \mathbb{R}^2$ denoting the CoM and stance foot positions.

\subsection{Jump Parameterization}
\label{sec:jump-parameterization}
To make the distribution of steering solutions easier to learn, we parameterize each jump by a low-dimensional action. A single jump is modeled as a transition between two consecutive takeoff states, consisting of a ballistic flight from the current takeoff state $\bm{x}$ followed by a stance phase driven by a parameterized GRF. The flight phase follows the closed-form ballistic dynamics in \eqref{eq:srb-dynamics} with zero GRF, and the touchdown state is obtained by evaluating this trajectory at the flight duration $T_{\mathrm{fl}}$.

During the stance phase, the GRF trajectory is parameterized by a cubic B\'ezier curve~\cite{ding2021hybrid}, and the foothold is specified by its horizontal position $p_x$, which determines the foot contact position $\bm{p}_{\mathrm{f}}^{\mathrm{st}} = [p_x, h(p_x)]^\top$ given the terrain height map $h(\cdot)$. Since the GRF must vanish at touchdown and takeoff, the curve is fully defined by its two intermediate control points $\bm{u}_{\mathrm{st}} = [\bm{\lambda}_0^\top, \bm{\lambda}_1^\top]^\top \in \mathbb{R}^4$:
\begin{equation}
       \bm{\lambda}(s) = 3(1-s)^2 s\,\bm{\lambda}_0 + 3(1-s)s^2\,\bm{\lambda}_1,
       \label{eq:bezier-grf}
\end{equation}
where $s = t / T_{\mathrm{st}} \in [0, 1]$ is the normalized stance time and $T_{\mathrm{st}}$ is the stance duration.
A single jump is therefore defined by the action vector $\bm{u} = [T_{\mathrm{fl}}, p_x, T_{\mathrm{st}}, \bm{u}_{\mathrm{st}}^\top]^\top \in \mathbb{R}^7$, and integrating the dynamics over both phases yields a closed-form state transition map $\mathcal{F}: \mathbb{R}^6 \times \mathbb{R}^7 \to \mathbb{R}^6$:
\begin{equation}
    \bm{x}^{+} = \mathcal{F}(\bm{x}, \bm{u}),
    \label{eq:jump-primitive}
\end{equation}
where $\bm{x}$ and $\bm{x}^{+}$ denote the current and subsequent takeoff states, respectively.

\subsection{Feasibility Constraints}
\label{sec:constraints}
Since the SRB model abstracts away the leg, a planned jump must satisfy additional constraints to be trackable by the full robot, including the friction cone, input limits, joint limits, and terrain collision. The joint-space constraints are imposed through a full leg kinematics model, following~\cite{zhuang2025kinodynamic}. Terrain collision is checked for both the pitched body and the foot, where the flight-phase foot trajectory is simplified to an interpolation between takeoff and touchdown with a prescribed apex height. We denote by $\mathcal{U}_{\mathrm{adm}}(\bm x)$ the set of admissible actions from state $\bm x$ whose jumps satisfy all of these constraints.

\section{FlowRRT}

FlowRRT plans a sequence of jumps on discrete terrain by growing a tree whose nodes are takeoff states connected by single jumps. Each growth step picks a target foothold and finds the feasible jumps from an existing node that can land there, which is the steering problem at the core of the method. This problem is formulated as finding the entire set of feasible jumps rather than one of them, and both an offline expert solver and its online distilled sampler are designed around this set-valued target. On top of this steering, several designs in the tree search improve the efficiency of steering queries. The remainder of this section details each of these components.


\subsection{Task Space Sampling}
\label{sec:task-space-sampling}
To improve sampling efficiency, we follow the task-space sampling of~\cite{shkolnik2009path}, where samples are drawn in a low-dimensional task space to bias the expansion while the tree itself grows in the full state space.
In our setting, the task space $\mathcal{P} \subset \mathbb{R}$ is the set of horizontal foothold positions on the terrain. Each iteration draws a foothold $p_{\mathrm{rand}} \in \mathcal{P}$ and expands the tree toward a new takeoff state whose reachable set contains $p_{\mathrm{rand}}$. The reachable set of a takeoff state $\bm x$ is defined as:
\begin{equation}
    \mathcal{R}(\bm{x})
    =
    \left\{
    p_x \;\middle|\;
    \exists\, t \ge 0:
    [p_x,h(p_x)]^\top
    -
    \bm{p}_{\mathrm{c}}(t)
    \in \mathcal{W}
    \right\},
    \label{eq:reachable-footholds}
\end{equation}
where $\bm{p}_{\mathrm c}(t)$ is the CoM position along the ballistic trajectory from $\bm{x}$ and $\mathcal W\subset\mathbb R^2$ is the leg workspace. Sampling in task space reduces the dimension of the sampling space and also preserves the multimodality of the steering problem, as a takeoff state can reach the same target foothold in several distinct ways, for example by planting the stance foot on different terrain segments.

\subsection{Set-Valued Steering Formulation}
For monoped jumping, the discrete terrain and the complicated constraints leave a narrow and disconnected feasible region for each steering query, so neither the optimal action nor a single feasible action is a reliable target for distillation. Both discard the alternative solutions that nearby queries may need, which may make the mapping from query to action discontinuous. The steering BVP is therefore formulated as a set-valued feasibility problem whose solution is the entire set of feasible actions.
Given the sampled target foothold $p_{\mathrm{rand}}$ and the takeoff state $\bm{x}_\mathrm{sel}$ of the selected node, the BVP is formulated as:
\begin{equation}
    \begin{aligned}
        \min_{\bm u} \quad & 0 \\
        \text{s.t.} \quad
        & \bm u \in \mathcal U_{\mathrm{adm}}(\bm x_\mathrm{sel}), 
        \quad p_\mathrm{rand} \in \mathcal R(\mathcal F(\bm x_\mathrm{sel},\bm u)).
    \end{aligned}
    \label{eq:steer-feasibility-problem}
\end{equation}
The solution set is denoted by $\mathcal{U}_{\mathrm{feas}}(\bm c)$, where the condition $\bm{c} = (\bm{x}_{\mathrm{sel}}, p_{\mathrm{rand}}, h(\cdot))$ encodes the query information.

\begin{algorithm}[t]
\caption{Multi-start MPPI-CEM expert for one query}
\label{alg:expert}
\begin{algorithmic}[1]
\Require query $\bm c$, action space $\mathcal U$, initial std $\bm\sigma_0$, restarts $R$, iterations $M$, batch $B$, elite ratio $\rho$, tolerance $\varepsilon$
\Ensure feasible action set $\mathcal D$
\State $\mathcal{D}\gets\emptyset$
\For{$r = 1,\dots,R$}
    \State $\bm{u}\sim\Call{Uniform}{\mathcal U}$,\;
           $\bm{\sigma}\gets\bm{\sigma}_0$
    \For{$m = 1,\dots,M$}
        \State $\bm{u}_b\sim\mathcal{N}(\bm u,\mathrm{diag}(\bm{\sigma}^2))$,\; $b = 1,\dots,B$
        \State $\{J_b, J^{\mathrm{vio}}_b\}\gets\Call{Rollout}{\bm{c},\{\bm{u}_b\}}$
        \State $\mathcal{D}\gets\mathcal{D}\cup\{\bm{u}_b : J^{\mathrm{vio}}_b\le\varepsilon\}$
        \State $(\bm{u},\bm{\sigma})\gets\Call{UpdateMPPICEM}{\bm{u},\bm{\sigma},\{\bm{u}_b,J_b\},\rho}$
    \EndFor
\EndFor
\State \Return $\mathcal{D}$
\end{algorithmic}
\end{algorithm}

\subsection{MPPI-based Expert Steering Solver}\label{sec:mppi-expert}
Considering the discrete footholds and the set-valued formulation, MPPI is applied here owing to its gradient-free updates and population-based search. Moreover, the action parameterization in Sec.~\ref{sec:jump-parameterization} makes each dynamics rollout a computationally cheap closed-form integration, which naturally fits the shooting nature of MPPI. As the constraints cannot be enforced in shooting, they are injected as soft penalties, and the resulting optimization problem reads
\begin{equation}
       \min_{\bm{u}} \quad
       J(\bm{u} \mid \bm c) = \Phi(\bm{u})
         + \underbrace{\sum_{j} w_j \max_{k} \big[g_j(\bm{x}_k, \bm{u} \mid \bm c)\big]_+}_{J^{\mathrm{vio}}(\bm{u} \mid \bm c)},
       \label{eq:expert-ocp}
\end{equation}
where $\bm{x}_k = \bm{x}(t_k)$, $k = 0,\dots,N-1$, are the checkpoints of the closed-form trajectory from $\bm x(0) = \bm{x}_{\mathrm{sel}}$, $g_j$ denote the constraint violations, and $\Phi$ is a quadratic GRF effort term. The effort term aims to provide a mild bias that keeps the sampling distribution moving across the feasible set rather than settling once it enters, since $J^{\mathrm{vio}}$ is flat inside the set.

Among MPPI variants, the MPPI-CEM optimizer of~\cite{hansen2022temporal} is adopted and run from multiple restarts, as summarized in Algorithm~\ref{alg:expert}, where \textsc{UpdateMPPICEM} moves the nominal action toward the softmax-weighted mean of the elite perturbations and blends the covariance toward their weighted variance. Multiple restarts are used to cover as many regions of $\mathcal U_{\mathrm{feas}}(\bm c)$ as possible, and every feasible sample met along the way is kept rather than only the cheapest. The cost of the restarts is modest, as all rollouts across restarts and batches are independent and run in parallel on the GPU~\cite{macklin2022warp}.

\subsection{CFM-based Generative Action Sampler}\label{sec:CFM}

\begin{figure}[t]
\centering
\includegraphics[width=0.42\textwidth]{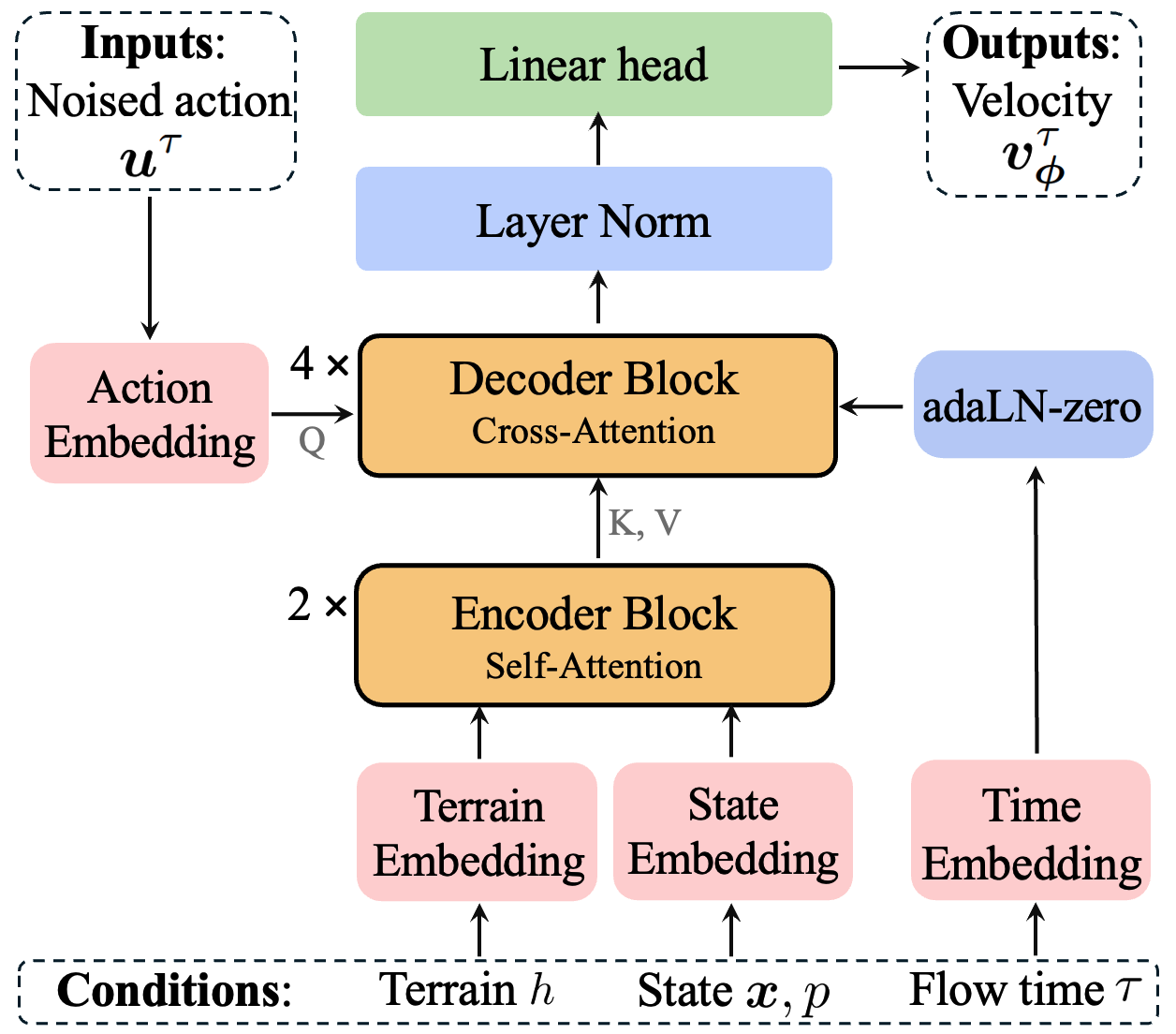}
\caption{Architecture of the conditional velocity field $\bm v_{\bm\phi}^{\tau}(\bm u^\tau;\bm c)$. The conditions are embedded into one state--target token and one token per cell of the local height map, each carrying its position relative to the target foothold and the takeoff foot, and encoded by self-attention blocks. The noised action $\bm u^\tau$ is embedded as a single token and refined by decoder blocks, where it provides the queries and the encoded context the keys and values. The flow time $\tau$ modulates the decoder through adaLN-Zero, and a linear head outputs the velocity.}
\label{fig:network-arch}
\end{figure}

Distilling the set-valued steering calls for a generative model rather than a regressor, which naturally covers a disconnected feasible set and yields many candidates per query. Among generative models, CFM is chosen for its fast sampling, and the target of the sampler is the uniform distribution $p^*(\bm u\mid\bm c)$ over $\mathcal{U}_{\mathrm{feas}}(\bm{c})$. Since the effort term in the cost biases the expert samples toward low-cost actions, the collected actions are post-processed to remove this bias, yielding the training distribution $p_{\mathrm{data}}(\bm u\mid\bm c)$ as an approximation of $p^*$.

Following the standard rectified-flow recipe~\cite{lipman2023flow,liu2023rectified}, a conditional velocity field $\bm v_{\bm\phi}^{\tau}(\bm u;\bm c)$ is trained to transport Gaussian noise $\bm u^0\sim\mathcal N(\bm 0,\bm I_7)$ to feasible actions $\bm u^1\sim p_{\mathrm{data}}(\cdot\mid\bm c)$ along the straight interpolant $\bm u^\tau=(1-\tau)\bm u^0+\tau\bm u^1$, $\tau\sim\mathcal U(0,1)$, by minimizing
\begin{equation}
\mathcal L_{\mathrm{CFM}}(\bm\phi)
=
\mathbb E\left[
\left\|
\bm v_{\bm\phi}^{\tau}(\bm u^\tau;\bm c)
-(\bm u^1-\bm u^0)
\right\|_2^2
\right].
\label{eq:flow-matching-loss}
\end{equation}

The conditional velocity field $\bm v_{\bm\phi}^{\tau}(\bm u;\bm c)$ is parameterized by the encoder--decoder Transformer in Fig.~\ref{fig:network-arch}. The terrain is tokenized per cell rather than flattened, so that attention can relate the action to the spatial layout of footholds around the robot, and the action is kept as a single token that queries this context.
At inference, a batch of Gaussian samples is transported in parallel through the learned flow with a few Euler steps. Since the learned distribution only approximates $p^*$, each candidate is rolled out and checked against the constraints of \eqref{eq:steer-feasibility-problem}, and only the feasible ones are returned for tree expansion.

\subsection{Reachability Estimator}\label{sec:reachability}

To avoid wasting steering queries on parents that cannot reach the sampled foothold, we follow reachability-guided sampling~\cite{shkolnik2009reachability} with a learned estimator. The feasible action rate of a query $\bm c$ is defined as the probability that a sample from the flow policy $p_{\bm\phi}(\bm u\mid\bm c)$ is feasible,
\begin{equation}
r_{\bm\phi}(\bm{c}) =
\mathop{\mathbb{P}}_{\bm{u}\sim p_{\bm\phi}(\cdot\mid\bm{c})}
\!\left[
    \bm{u}\in\mathcal{U}_{\mathrm{feas}}(\bm{c})
\right],
\label{eq:feasible-action-rate}
\end{equation}
and a regressor $\hat r_{\bm\psi}(\bm c)$ is trained to predict it from randomly sampled queries, each labeled by the empirical fraction of feasible candidates among a batch of rollouts from the flow policy as a Monte Carlo estimate of $r_{\bm\phi}$. Parents whose predicted rate falls below a threshold are rejected before steering. 

\begin{algorithm}[t]
\caption{FlowRRT}
\label{alg:flow-rrt}
\begin{algorithmic}[1]
\Require terrain $h$, root state $\bm x_{\mathrm{root}}$, goal region $\mathcal{P}_{\mathrm{goal}}$,
         flow policy $p_{\bm\phi}$, reachability estimator $\hat r_{\bm\psi}$,
         batch $B$, steer budget $K$, tolerance $\varepsilon$
\Ensure tree $\mathcal T$ and solution path
\State $\mathcal T=(\mathcal V,\mathcal V_{\mathrm{active}},\mathcal E)\gets(\{\bm x_{\mathrm{root}}\},\{\bm x_{\mathrm{root}}\},\emptyset)$
\For{$i = 1,\dots,K$}
    \Repeat
        \State $p_{\mathrm{rand}}\gets\Call{RandomSample}{h,\mathcal{P}_{\mathrm{goal}}}$
        \State $\mathcal V_{\mathrm{feas}}\gets\Call{ReachFilter}{\mathcal V_{\mathrm{active}},p_{\mathrm{rand}},h,\hat r_{\bm\psi}}$
    \Until{$\mathcal V_{\mathrm{feas}}\neq\emptyset$}
    \State $\bm{x}_{\mathrm{sel}}\gets\Call{SelectParent}{\mathcal V_{\mathrm{feas}},p_{\mathrm{rand}}}$
    \State $\mathcal U_{\mathrm{feas}}\gets\Call{Steer}{\bm x_{\mathrm{sel}},p_{\mathrm{rand}},h,p_{\bm\phi},B,\varepsilon}$
    \State $\mathcal T\gets\Call{Prune}{\mathcal T,\bm x_{\mathrm{sel}},\mathcal U_{\mathrm{feas}}}$
    \State $\mathcal U^{\star}\gets\Call{MultiModalExpand}{\bm x_{\mathrm{sel}},\mathcal U_{\mathrm{feas}}}$
    \State $\mathcal X_{\mathrm{new}}\gets\Call{AddChildren}{\mathcal T,\bm{x}_{\mathrm{sel}},\mathcal U^{\star}}$
    \If{$\exists\,\bm x\in\mathcal X_{\mathrm{new}}:\ \mathcal R(\bm x)\cap\mathcal P_{\mathrm{goal}}\ne\emptyset$}
        \State \Return $\mathcal T$ and $\Call{ExtractPath}{\mathcal T,\bm x}$
    \EndIf
\EndFor
\State \Return $\mathcal T$ and failure
\end{algorithmic}
\end{algorithm}

\subsection{FlowRRT}\label{sec:flow-rrt}

\begin{figure}[!t]
    \centering
    \includegraphics[width=0.48\textwidth]{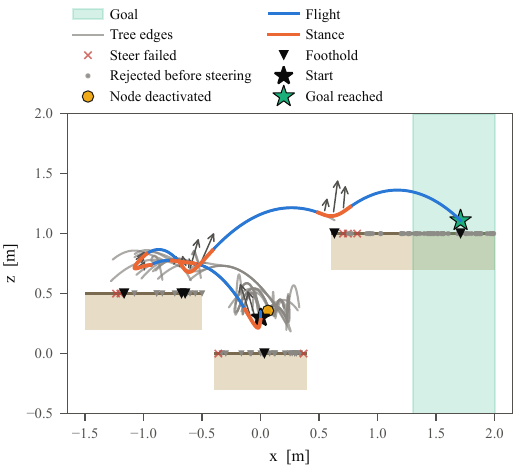}
    \caption{One FlowRRT search, following Algorithm~\ref{alg:flow-rrt}. A target foothold is sampled, and parents predicted by the reachability estimator to be unable to reach it are rejected before steering. The selected parent is steered toward the target, which either yields new tree edges or fails, and a node that fails repeatedly is deactivated. Once a node reaches the goal region, the path is extracted, here jumping backward onto the lower platform before clearing the gap.}
    \label{fig:rrt-process}
\end{figure}

Algorithm~\ref{alg:flow-rrt} summarizes FlowRRT. Since each steering query is the dominant cost and returns a whole feasible set rather than one action, the tree search is designed around how to avoid unnecessary steers and how to fully leverage the feasible set from a successful one. The remainder of this section details the corresponding subroutines.

To spend a query only where it is likely to succeed, \textsc{ReachFilter} keeps the active nodes whose predicted rate $\hat r_{\bm\psi}$ for the sampled foothold exceeds a threshold. Even so, some nodes repeatedly fail to yield a feasible extension, especially when the estimator overestimates their reachability. 
\textsc{Prune} therefore tracks for each node how often it has been selected as a parent, $n_{\mathrm{sel}}(\bm x)$, and how many of those steers returned an empty $\mathcal U_{\mathrm{feas}}$, $n_{\mathrm{fail}}(\bm x)$, and deactivates the node once $n_{\mathrm{sel}}(\bm x)\ge n_{\min}$ and $n_{\mathrm{fail}}(\bm x)/n_{\mathrm{sel}}(\bm x)>\eta_{\mathrm{fail}}$.
Deactivated nodes leave $\mathcal V_{\mathrm{active}}$ but remain in $\mathcal V$, so that paths through them stay valid.

Instead of adding one child per steer, \textsc{MultiModalExpand} adds several by clustering the feasible rollouts on their toe-off state and stance foothold, so that the diverse feasible actions returned by the sampler are fully leveraged. The clusters are formed by complete-linkage hierarchical clustering cut at a fixed distance threshold, and the cheapest action of each cluster is returned as $\mathcal U^\star$.

\begin{figure*}[tb]
    \centering
    \includegraphics[width=0.9\textwidth]{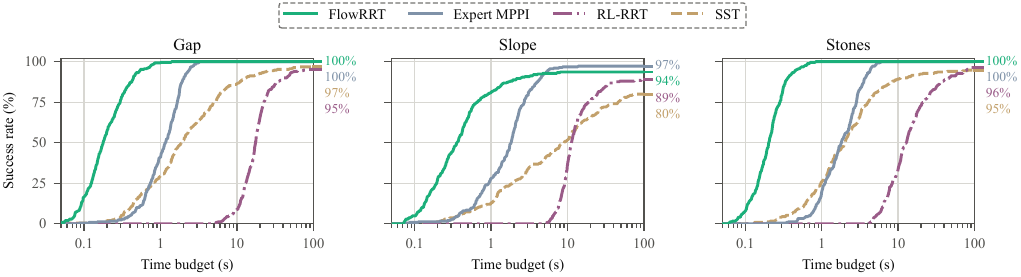}
    \caption{Success rate versus planning time for FlowRRT, RL-RRT~\cite{rlrrt2019}, and SST~\cite{li2016asymptotically} baselines, and the Expert MPPI reference across terrain classes.  For each terrain class, all planners are evaluated on the same 50 randomly generated terrain instances spanning $x \in [-1, 6]$~m, with the robot starting at $x = 0$ and the goal region defined as $\mathcal{P}_{\mathrm{goal}} = [5, 6]$~m. Each terrain is evaluated with five planner seeds, yielding 250 runs per method per terrain class. At time $t$, the success rate is the fraction of these runs that have found a valid solution within $t$ seconds, and the value at the right edge of each panel is the success rate at the full budget of $100$~s.}
    \label{fig:benchmark}
\end{figure*}

The remaining subroutines follow standard practice. \textsc{SelectParent} draws the parent at random among the top-$k$ nodes closest to $p_{\mathrm{rand}}$ under a weighted Euclidean distance, which encourages exploration. \textsc{Steer} samples a batch of actions from $p_{\bm\phi}$, rolls them out, and returns the feasible subset $\mathcal U_{\mathrm{feas}}$, and \textsc{AddChildren} inserts the new takeoff states into $\mathcal V$ and $\mathcal V_{\mathrm{active}}$ with their edges.

\section{Results}

\subsection{Setup}

\subsubsection{Robot}
The validation platform is a custom planar monoped weighing $1.97$~kg, with a two-link leg of $0.22$~m per link. The feasibility constraints use a friction coefficient of $\mu = 0.8$, a torque limit of $20$~N$\cdot$m at the hip and knee, and the joint ranges of the hardware.

\subsubsection{Computation}
All experiments were conducted on a single workstation with an Intel Core i7-13700 CPU and an NVIDIA RTX~5070 GPU, using PyTorch~2.12 and CUDA~13.0. All SRB rollouts, both inside the multi-start MPPI-CEM expert and in the feasibility checks during planning, were implemented in NVIDIA Warp~\cite{macklin2022warp}, and both rollouts and policy inference ran in FP32, with CUDA Graphs used for inference during planning.

\subsubsection{Terrain}
Terrains are randomly generated as a sequence of variable-width blocks, each of which is either a ramp platform between its two end heights or a gap, with all parameters drawn uniformly within their bounds. The initial takeoff state and foothold are fixed with respect to the terrain underneath. Three terrain types stress different aspects of the problem. \textit{Gap} has wide level platforms at varying heights separated by unlandable gaps, which disconnects the feasible footholds. \textit{Slope} lets the same platforms lean, which tightens the friction cone and pitch constraints at landing. \textit{Stones} narrows the platforms to sparse stepping stones, which demands precise landing.

\subsubsection{Expert Data Generation}
The expert of Algorithm~\ref{alg:expert} is run with the settings in Table~\ref{tab:mppi-cem}. The feasible sets are deduplicated in the normalized action space with a radius of $0.1$, queries left with fewer than $2000$ actions are re-solved with $10$ additional restarts, and about $5\%$ of the queries are held out for evaluation.

\begin{table}[t]
\centering
\caption{Multi-start MPPI-CEM expert algorithm settings}
\label{tab:mppi-cem}
\small
\setlength{\tabcolsep}{4pt}
\begin{tabular}{@{}llr@{}}
\toprule
$R$              & restarts per query                            & $40$      \\
$M$              & iterations per restart                        & $20$      \\
$B$              & rollouts per iteration                        & $1024$    \\
$\rho$           & elite fraction                                & $0.5$     \\
$\beta$          & softmax temperature                           & $0.5$     \\
$\alpha$         & nominal step size                             & $0.8$     \\
$\gamma$         & covariance adaptation rate                    & $0.5$     \\
$\sigma_{\min}$  & floor on the adapted standard deviation       & $10^{-3}$ \\
$\varepsilon$    & feasibility tolerance on $J^{\mathrm{vio}}$   & $10^{-3}$ \\
\bottomrule
\end{tabular}
\end{table}

\subsubsection{Flow Matching Steer Policy}
The velocity-field network has two encoder and four decoder blocks with hidden dimension $128$ and four attention heads, totaling $1.70$M parameters. The local height map covers $[-1,3]$~m in the takeoff frame at $5$~cm resolution, and actions and state features are standardized while heights remain in meters. The model is trained with AdamW at a learning rate of $3\times10^{-4}$ and batch size $512$, and inference uses $10$ Euler steps.

\subsubsection{Planner Configuration}
FlowRRT draws $B = 1024$ candidate actions per steer and selects the parent uniformly among the top $k = 5$ nodes ranked by a Euclidean distance to the sampled foothold with weights $(1, 3)$ on the horizontal and vertical components. The goal bias is $0.2$, the reachability threshold is $\hat r_{\bm\psi}(\bm c) \ge 0.01$, and the clustering distance threshold for multi-modal expansion is $2.0$. A node is pruned from the active set after $n_{\min} = 5$ parent trials once its failure ratio exceeds $\eta_{\mathrm{fail}} = 0.8$, and the steer budget is $K = 1000$ for the benchmark and ablations.

\subsubsection{Baselines}
\label{sec:baselines}

FlowRRT is compared against representative planners based on forward propagation without steering and on policy-based steering, together with the expert it is distilled from. SST~\cite{li2016asymptotically} propagates the same SRB model with controls drawn from a Gaussian fitted to the expert data, since a uniform prior solved none of the instances. RL-RRT~\cite{rlrrt2019} steers with a goal-reaching policy trained by proximal policy optimization (PPO)~\cite{schulman2017proximal} on the full-order hopper, run in closed loop for up to $4$~s per steer. Expert MPPI steers with the multi-start MPPI-CEM solver inside the same framework, so that its gap to FlowRRT measures the cost of distillation.

\begin{figure*}[tb]
       \centering
       \includegraphics[width=1\linewidth]{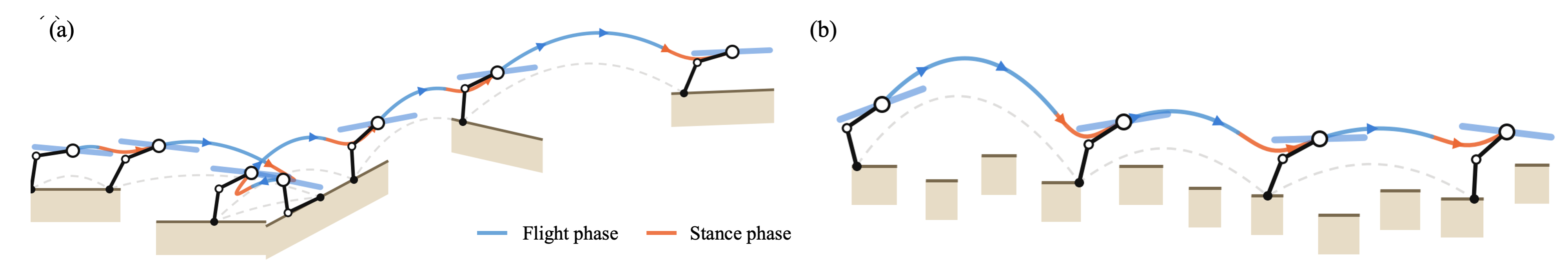}
    \caption{Planned motions on two terrains. Robots mark the contacts of the returned path, drawn at the end of each stance. The solid curve is the CoM path, blue in flight and orange in stance, the dashed curve is the foot path, and arrowheads mark the direction of takeoff velocity. (a) \textit{Slope} terrain. Arriving too fast to climb after the third jump, the robot steps back onto the flat platform and returns to the slope with a slower approach and a higher foothold for the climb. (b) \textit{Stones} terrain. The path crosses stepping stones no wider than $0.24$~m with one contact per stone.}
       \label{fig:classical-terrain}
\end{figure*}


\subsection{Benchmark}

Fig.~\ref{fig:benchmark} reports the success rate against the planning time budget on the three terrain classes. Compared with the two baselines, FlowRRT attains the highest success rate over the whole range of budgets, and the margin is largest under tight budgets. Within one second, FlowRRT solves $99.2\%$ of the runs on \textit{Gap}, $80.8\%$ on \textit{Slope}, and all runs on \textit{Stones}, whereas SST reaches less than half of these rates and RL-RRT solves none. Fig.~\ref{fig:classical-terrain}(b) shows a representative path on \textit{Stones}, where the admissible footholds are narrow.
The two baselines fall behind for different reasons. SST propagates random controls, which rarely yield feasible jumps for this underactuated system, so its success rate climbs slowly with the budget, pointing to the necessity of a steering BVP solver. RL-RRT, in contrast, steers reliably but at a high computational cost, since every extension simulates the policy in closed loop for up to $4$~s of motion. Formulating the steering BVP directly over the open-loop action trajectory of an entire jump instead yields a far faster query than such policy-based steering.

Against the expert it is distilled from, FlowRRT shows that distillation is highly effective. Within one second it solves at least twice as many runs as Expert MPPI on every terrain class. With a larger budget, however, Expert MPPI attains a higher final success rate on \textit{Slope}, indicating that the learned steering distribution does not fully capture the feasible actions that the expert discovers in the hardest cases. Distillation thus trades a small loss in final success rate on the most demanding terrain for a large gain in online steering speed.

\subsection{Planning on Challenging Terrain}

The planned motions exhibit the terrain-exploiting behavior that motivates this work, in which the planner leaves the direct route and uses intermediate footholds to reset its takeoff state. Fig.~\ref{fig:rrt-process} and Fig.~\ref{fig:classical-terrain}(a) show two such detours, the first to gain height for a high platform and the second because the friction cone forbids climbing the incline at the arriving speed. Such detours are also where explicit planning differs from end-to-end perceptive locomotion, in which a goal-conditioned policy plans footholds implicitly. In our observation, such a policy trained with PPO on randomly generated terrains of the same three classes traverses them through forward jumps but does not discover the backward jump in Fig.~\ref{fig:rrt-process}, since the goal reward is sparse while progress rewards penalize moving away from the goal.


\subsection{Ablation Study}
\begin{figure}[t]
    \centering
    \includegraphics[width=0.48\textwidth]{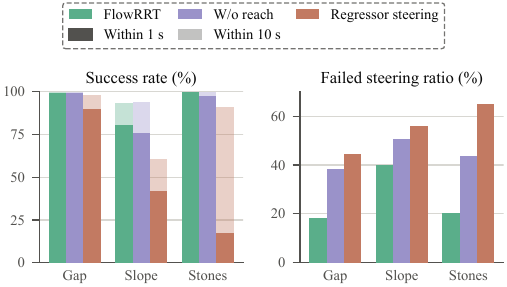}
    \caption{Ablation of FlowRRT on the same runs as Fig.~\ref{fig:benchmark}. W/o reach removes the reachability estimator, and regressor steering replaces the flow sampler with a regressor, with everything else kept the same. Left, success rate within 1~s in solid color and within 10~s in light color. Right, failed steering ratio, the share of steering calls over all runs that yield no feasible child node.}
    \label{fig:ablation}
\end{figure}

We ablate the two learned components of the planner in Fig.~\ref{fig:ablation}. Removing the reachability estimator tests how much the planner gains from skipping unreachable queries before steering, and replacing the flow policy with a regressor of the lowest-cost action tests how much the generative formulation itself contributes. The regressor shares the backbone, training data, and estimator training procedure with FlowRRT, and Gaussian perturbations are added around its single prediction to give it the same number of candidates as the flow policy.

Both components contribute by reducing the failed steering ratio and thereby speeding up planning. Removing the reachability estimator roughly doubles the failed steering ratio on \textit{Gap} and \textit{Stones}. Although the success rate barely drops, the median planning time increases by $60$--$67\%$. Replacing the flow policy with regressor steering costs both speed and success.
Within 1~s, it solves only $42.0\%$ of the \textit{Slope} runs and $17.2\%$ of the \textit{Stones} runs, compared with $80.8\%$ and $100\%$ for FlowRRT. On the most demanding \textit{Slope}, its success rate stays at $60.4\%$ even within 10~s, and on \textit{Stones} its failed steering ratio rises to $65.0\%$, pushing the median planning time from $0.20$ to $1.75$~s. This comparison supports our key insight that predicting the feasible actions as a distribution substantially outperforms regressing a single action, especially when the feasible set is narrow, as on \textit{Slope} and \textit{Stones}.

\subsection{Tracking Experiment}
\begin{figure}[t]
    \centering
    \includegraphics[width=0.48\textwidth]{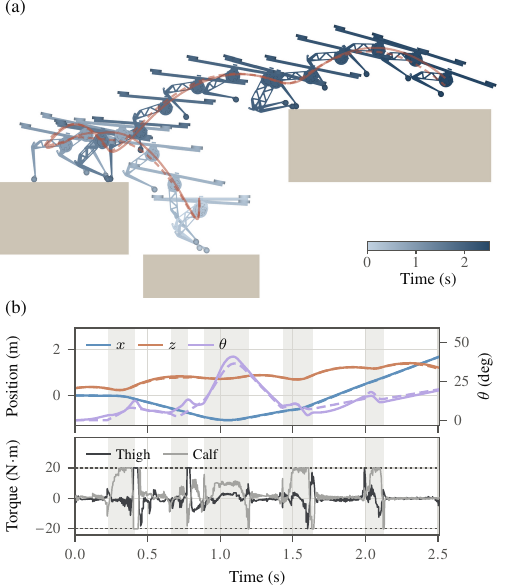}
    \caption{Tracking experiment using a learned tracking policy.
(a) Simulation snapshots, with solid and dashed lines indicating the simulated and FlowRRT planned CoM trajectories, respectively.
(b) State and action trajectories during execution.}
    \label{fig:tracking}
\end{figure}

To evaluate whether the planned trajectory can be executed under full-order dynamics, we select the terrain and trajectory shown in Fig.~\ref{fig:rrt-process} as a representative tracking task. We train a tracking policy on the FlowRRT reference with PPO in Isaac Lab, building on the motion-tracking formulation of BeyondMimic~\cite{liao2026beyondmimic} with additional foothold and terrain-aware terms, and evaluate it in MuJoCo for sim-to-sim validation, shown in Fig.~\ref{fig:tracking}. The robot completes all five planned landings in each of $16$ evaluation runs, with base tracking RMSEs of $1.48$~cm, $2.07$~cm, and $2.37^\circ$ for horizontal position, vertical position, and pitch, respectively. The mean absolute foothold error is $0.68$~cm, and the largest per-landing error averaged over the runs is $1.37$~cm. These results demonstrate accurate execution of the representative planned motion under full-order dynamics.


\section{Conclusion and Future Work}
This paper presented FlowRRT, a kinodynamic RRT planner for agile monoped hopping on discrete terrain, and demonstrated that learning steering solutions as a distribution rather than a single action is an effective way to bring steering-based RRT to legged robots. The proposed framework poses each steer as a set-valued BVP over the open-loop action trajectory of a single jump, collects its feasible solutions with a multi-start MPPI expert, and distills them into a CFM sampler. Across all three terrain classes, FlowRRT attained the highest success rate over the whole range of planning budgets, with the largest margin under tight budgets, and a planned trajectory was tracked accurately under full-order dynamics, verifying that the SRB-based plans are executable by the full robot.

Regarding the limitations, the evaluation is confined to simulation. Hardware experiments on a physical monoped are the immediate next step, followed by the extension to bipedal and humanoid jumping.




\bibliographystyle{IEEEtran}
\bibliography{refs}

\end{document}